\documentclass[preprint,12pt,authoryear]{elsarticle}
\usepackage[a4paper,margin=25mm]{geometry}
\usepackage[T1]{fontenc}
\usepackage[utf8]{inputenc}
\usepackage{lmodern}
\usepackage[final]{microtype}
\usepackage{amsmath,amssymb,amsfonts}
\usepackage{graphicx}
\usepackage{textcomp}
\usepackage{booktabs,tabularx,array}
\usepackage{placeins}
\usepackage{xurl}
\usepackage[hidelinks]{hyperref}
\hypersetup{pdftitle={HGASurv: Hierarchical Granularity-Aware Multimodal Integration for Survival Prediction},pdfauthor={Peike Yu, Wuchao Liu, Jihong Guan, Shuigeng Zhou},pdfsubject={},pdfkeywords={}}
\biboptions{authoryear,round,semicolon,sort}
\newcommand{\rev}[1]{#1}
\AddToHook{env/table/begin}{\setlength{\belowcaptionskip}{8pt}}
\renewcommand{\thefnote}{\ensuremath{\dagger}}
\makeatletter
\def\ps@pprintTitle{\let\@oddhead\@empty\let\@evenhead\@empty
  \def\@oddfoot{\hfil\thepage\hfil}\let\@evenfoot\@oddfoot}
\makeatother
\begin{document}
\raggedbottom
\begin{frontmatter}
\title{HGASurv: Hierarchical Granularity-Aware Multimodal Integration for Survival Prediction}

\author[tongji]{Peike Yu\fnref{equal}}
\ead{2451951@tongji.edu.cn}
\author[tongji]{Wuchao Liu\fnref{equal}}
\ead{wuchaoliu@tongji.edu.cn}
\author[tongji]{Jihong Guan\corref{corresponding}}
\ead{jhguan@tongji.edu.cn}
\author[fudan]{Shuigeng Zhou\corref{corresponding}}
\ead{sgzhou@fudan.edu.cn}
\address[tongji]{School of Computer Science and Technology, Tongji University,
1239 Siping Road, Shanghai 200092, China}
\address[fudan]{College of Computer Science and Artificial Intelligence,
Fudan University, 2005 Songhu Road, Shanghai 200438, China}
\fntext[equal]{Peike Yu and Wuchao Liu contributed equally to this work and share first authorship.}
\cortext[corresponding]{Corresponding authors: Jihong Guan and Shuigeng Zhou.}

\begin{abstract}
\linespread{1}\selectfont
\rev{Multimodal survival prediction jointly analyzes
  histopathological whole-slide images (WSIs), genomic profiles, and clinical variables to
  stratify patients by risk. However, existing approaches often inadequately account for
  granularity differences across modalities, particularly in spatial organization and
  feature dimensionality. Within pathology, aggregating WSI patch features without
  preserving their spatial arrangement can obscure tissue architecture; across modalities,
  directly concatenating low-dimensional clinical variables with high-dimensional visual
  and genomic features may limit the contribution of clinical information to prediction.
  To address these challenges, we propose HGASurv, which introduces a hierarchical WSI
  encoding module and a clinical-guided graph propagation module for granularity-aware
  multimodal integration.} 
  \rev{The hierarchical WSI encoding module
  maps WSI patch features onto a spatial grid using their original coordinates and applies
  parallel convolutions to construct representations at patch, local-neighborhood, and
  regional scales. Genomic signature embeddings then query these scale-specific WSI
  representations through cross-modal co-attention, producing signature-conditioned
  morphological features. The clinical-guided graph propagation module uses clinical
  variables to define patient-similarity edges and propagates fused image--genomic node
  embeddings with graph attention, integrating clinical information through topology
  rather than feature concatenation.} 
  \rev{Across three TCGA cohorts,
  HGASurv achieved the strongest overall C-index performance, with ablation studies
  supporting the complementary contributions of the hierarchical WSI encoding and
  clinical-guided graph propagation modules.} 
  \rev{By introducing hierarchical WSI encoding and clinical-guided graph propagation,
  HGASurv unifies multi-scale pathology representation with topology-based clinical
  integration in a hierarchical granularity-aware framework. This design shows that
  matching representation and integration mechanisms to modality-specific granularity
  improves multimodal survival prediction.\looseness=-1}
\end{abstract}
\end{frontmatter}

\section{Introduction}
Survival analysis is a fundamental statistical framework in prognostic
research by accommodating censored data to simultaneously model the timing and occurrence
of clinical events \citep{kaplan1958,cox1972}. By stratifying patients into risk groups and
estimating survival probabilities over the follow-up period, survival prediction supports
treatment planning and informs patients about the likely course of disease
\citep{giordano}. Driven by recent advances in deep learning, researchers have shifted
their focus from single-modality analysis to integrating richer multimodal information,
thus achieving superior prognostic performance \citep{wiegrebe}. \rev{Histopathological
whole-slide images (WSIs) provide spatial information about tissue morphology and
architecture; genomic sequencing profiles describe the molecular state; and clinical
variables summarize patient- and disease-level status.} However, these diverse modalities
associated with patient survival outcomes exhibit substantial heterogeneity in information
granularity, particularly in spatial organization and feature dimensionality. Tissue
morphology in histopathological WSIs is spatially organized across multiple scales, and a
single WSI can measure up to $100{,}000 \times 100{,}000$ pixels \citep{jaume2024}. Genomic
profiles may contain tens of thousands of variables, whereas clinical data typically
comprise fewer than 100 variables. Such profound heterogeneity in information granularity
renders the integration of multimodal data for survival prediction highly challenging,
thereby demanding tailored model architectures.

A series of deep learning-based models has emerged for multimodal data fusion in survival
prediction. Most existing approaches, such as MCAT \citep{chen2021}, \rev{SurvPath}
\citep{jaume2024}, 
\rev{SurMoE} \citep{zhang2025surmoe} and DeReF \citep{wang2025}, focus on
integrating histopathological images with genomic profiles, while several works such as
PG-TFNet \citep{lv2021} and SAMMS \citep{zhu2023} further integrate clinical variables as an
additional modality. 
\rev{Although existing models have demonstrated promising performance
in survival prediction, certain key challenges still remain unaddressed.} On the one hand,
regarding large-scale histopathological images, the prevailing paradigm involves dividing
WSIs into small patches and extracting patch-level visual representations for downstream
fusion. Aggregating these patch-level features without preserving their spatial
arrangement can obscure the global tissue architecture, thereby hindering the model from
discerning pathological morphological patterns. On the other hand, regarding the fusion of
low-dimensional clinical variables, existing methods generally concatenate clinical
variables with high-dimensional features learned from other modalities before the final
survival head. Considering the substantial dimensionality disparity, this concatenation
strategy risks rendering clinical variables diluted or suppressed by dominant visual and
genomic features, which can limit their contribution to survival prediction.

To address the above issues, we propose a \textbf{H}ierarchical
\textbf{G}ranularity-\textbf{A}ware multimodal integration framework for \textbf{Surv}ival
prediction (\textbf{HGASurv}). 
\rev{To preserve WSI spatial organization, HGASurv
introduces a hierarchical WSI encoding module. Patch embeddings are arranged on a spatial
grid according to their original coordinates and processed by parallel convolutions to
derive representations at patch, local-neighborhood, and regional scales. Following the
genomic-signature grouping in MCAT \citep{chen2021}, six signature embeddings query each
scale-specific WSI representation through cross-modal co-attention. The resulting
pathology and genomic representations are fused into an image--genomic embedding. To
incorporate clinical variables without direct feature concatenation, HGASurv introduces a
clinical-guided graph propagation module. Clinical similarity defines an undirected
patient graph, over which a graph attention network propagates the fused image--genomic
embeddings. Clinical variables thus influence prediction through graph topology rather
than being appended to the image--genomic node features. Together, these components form a
hierarchy from intramodal multi-scale WSI encoding, through within-patient
pathology--genomics fusion, to clinically guided inter-patient propagation. This hierarchy
aligns each representation and integration mechanism with the information granularity of
the corresponding modality.} The main contributions of our work are as
follows.\looseness=-1
\begin{itemize}
  \item \rev{We introduce a hierarchical WSI encoding module that reconstructs patch
  embeddings on a spatial grid and derives multi-scale pathology representations at patch,
  local-neighborhood, and regional scales.}
  \item \rev{We introduce a clinical-guided graph propagation module that uses clinical
  variables to define patient-similarity topology and propagates fused image--genomic
  embeddings with graph attention, integrating clinical information through topology
  rather than feature concatenation.}
  \item \rev{We evaluate HGASurv on TCGA-GBMLGG, TCGA-BLCA, and TCGA-UCEC, achieving
  C-indices of $0.854 \pm 0.023$, $0.692 \pm 0.032$, and $0.721 \pm 0.098$, respectively
  (mean $\pm$ SD over five folds), and the strongest overall C-index performance across
  the three cohorts. Ablation studies support the complementary contributions of the
  hierarchical WSI encoding and clinical-guided graph propagation modules.}
\end{itemize}

\section{Related Work}
\subsection{\rev{Survival Prediction}}
\rev{Classical survival prediction commonly relies on the Cox proportional hazards
model~\citep{cox1972} and its penalized variants~\citep{lasso}, which relate covariates to
risk under the proportional-hazards assumption. Random Survival Forests~\citep{random}
capture nonlinear effects nonparametrically.} Neural models extend these foundations:
DeepSurv~\citep{katzman2018} replaces the Cox linear predictor with a multilayer perceptron
(MLP), whereas DeepHit~\citep{lee2018} directly models discrete-time event distributions
without the proportional-hazards assumption. Dynamic-DeepHit further handles longitudinal
data with competing risks \citep{lee2020ddh}. Although effective for structured covariates,
these methods are primarily unimodal and do not explicitly model interactions among
pathology, genomics, and clinical variables.

\subsection{\rev{Pathology--Genomics Multimodal Survival Prediction}}
\rev{Existing pathology–genomics survival methods explore different strategies for
integrating histological and genomic information. MCAT~\citep{chen2021} allows functional
genomic-signature embeddings to query histology features, whereas
SurvPath~\citep{jaume2024} models dense interactions between biological-pathway and
histology-patch tokens. MOTCat~\citep{xu2023} uses optimal transport to align histology
patches with genomic embeddings under global structure consistency.
SurMoE~\citep{zhang2025surmoe} employs dynamic expert routing to balance heterogeneous
inputs, while DeReF~\citep{wang2025} decouples shared, specific, and exclusive features
before reorganizing them for MoE fusion. \citet{liu2023vacol} investigate
view-aware collaborative learning for joint survival prediction and subgroup
identification.  However, these methods primarily focus on cross-modal interaction or
alignment, while explicit modeling of WSI spatial organization across multiple
morphological scales remains less explored.}

\subsection{\rev{Clinical Integration and Patient Relational Modeling}}
\rev{Compared with pathology and genomic data, structured clinical variables have received
less attention as a distinct modality in multimodal survival models. 
PG-TFNet~\citep{lv2021} and SAMMS~\citep{zhu2023} concatenate clinical variables with
learned pathology--genomics representations before the survival head.
ModalTune~\citep{ramanathan2025modaltune} instead uses a lightweight Modal Adapter to
inject auxiliary modalities into a frozen slide-level foundation model. \citet{sucre2025} compare early fusion with predictor-level late fusion, with the
latter combining outputs from independently trained unimodal models.
SurvPGC~\citep{hou2026survpgc} encodes clinical information as text embeddings and fuses
them with pathology and genomic representations through cross-attention. These approaches integrate clinical information within individual patients rather than using clinical similarity to guide cross-patient information exchange.}\par\rev{Patient-level graph modeling provides a
relational alternative. Patient-similarity graphs support information exchange among
related cases~\citep{parisot2018}, while graph attention networks (GATs) learn
neighbor-specific aggregation weights~\citep{velickovic2018}.
Cox--Sage~\citep{mao2025coxsage} exemplifies this strategy by constructing a patient graph
from heterogeneous clinical attributes, using gene-expression profiles as node features,
and applying GraphSAGE-based proportional-hazards modeling. However, Cox--Sage propagates
gene-expression representations and does not explicitly model multi-scale
pathology--genomics interactions. HGASurv combines multi-scale pathology–genomics fusion within patients with graph propagation across patients linked by clinical similarity.}

\begin{figure}[!htbp]
  \centering
  \includegraphics[width=\textwidth]{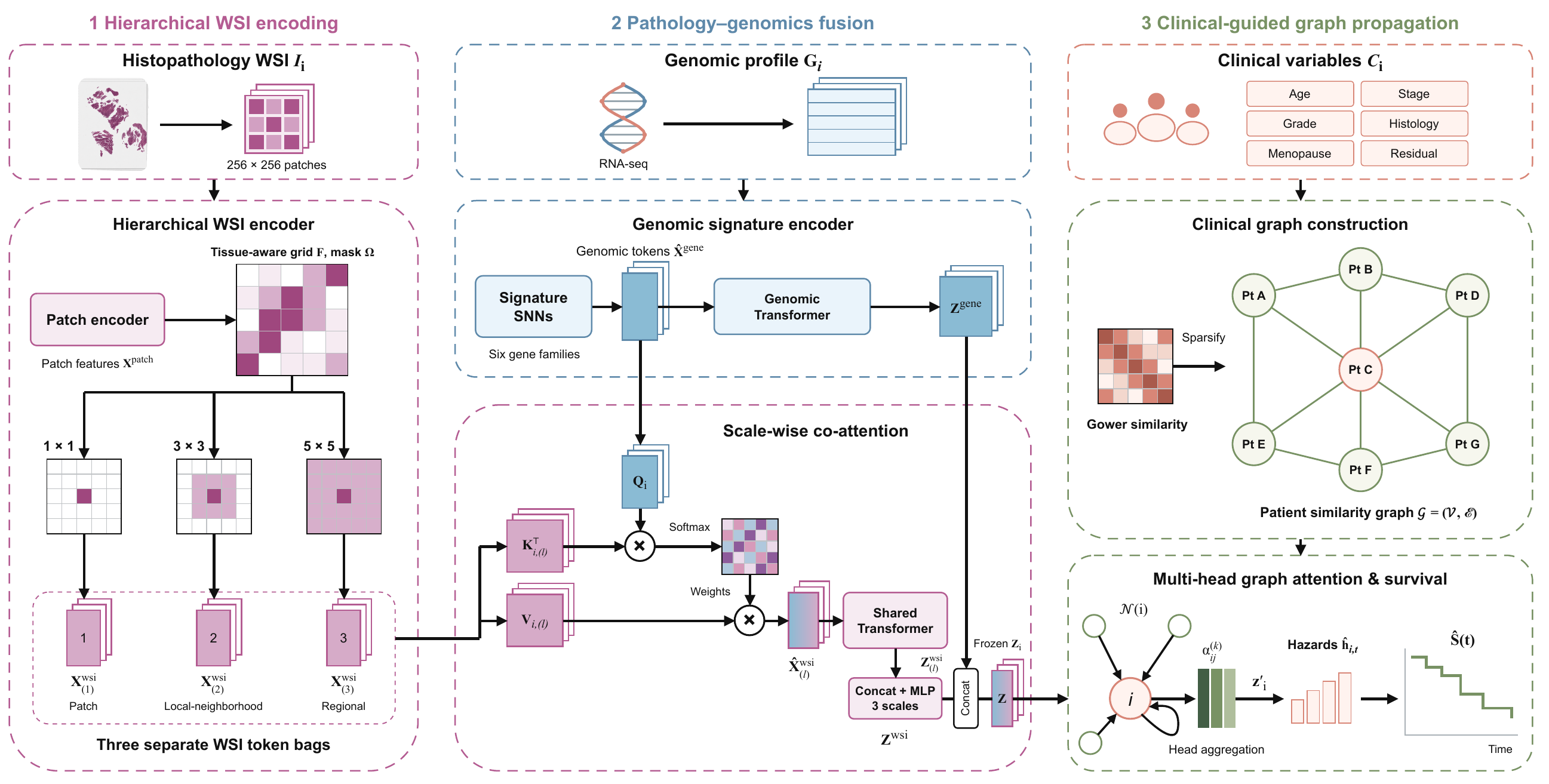}
  \caption{\rev{Overview of HGASurv. (1) Hierarchical whole-slide image (WSI) encoding
  preserves patch spatial organization and represents morphology at patch,
  local-neighborhood, and regional scales. (2) Genomic signatures guide scale-wise
  pathology aggregation through co-attention, yielding the fused image--genomic embedding
  $\mathbf{Z}_i$. (3) Clinical variables define patient relationships rather than node
  features; graph attention propagates $\mathbf{Z}_i$ over this topology to obtain
  $\mathbf{z}'_i$ for discrete-time survival prediction.}}
  \label{fig:framework}
\end{figure}

\section{Methods}
\label{sec:method}

\subsection{Overview of HGASurv}

\rev{Fig.~\ref{fig:framework} summarizes the HGASurv architecture, comprising hierarchical
WSI encoding, scale-wise pathology--genomics fusion, and clinical-guided graph propagation
for survival prediction.} We formulate multimodal survival prediction as discrete-time
hazard estimation. We consider a patient cohort $\mathcal{D} = \{(\mathbf{I}_i,
\mathbf{G}_i, \mathbf{C}_i, t_i, \delta_i)\}_{i=1}^N$, where $\mathbf{I}_i$ denotes the
WSI of patient $i$, $\mathbf{G}_i \in \mathbb{R}^{D_g}$ contains the genomic variables,
$\mathbf{C}_i$ contains $D_c$ mixed-type clinical variables, $t_i \in \mathbb{R}_{\geq 0}$
is the observed survival time, and $\delta_i \in \{0,1\}$ is the event indicator
($\delta_i=1$ for an observed event and $\delta_i=0$ for right-censoring). We discretize
survival time into $T$ non-overlapping intervals $\{\mathcal{B}_t\}_{t=1}^T$. HGASurv
estimates a discrete hazard probability $\hat{h}_{i,t} \in [0,1]$ for each patient $i$ and
interval $t$. The survival function $\hat{S}_i(t) = \prod_{u=1}^{t}(1-\hat{h}_{i,u})$
represents the probability that patient $i$ survives beyond the $t$-th time interval.

\rev{Hierarchical WSI encoding and scale-wise pathology--genomics fusion first produce a
patient-specific image--genomic embedding $\mathbf{Z}_i$. Clinical-guided graph
propagation then refines these embeddings over the patient graph to obtain
$\mathbf{z}'_i$, which is passed to the survival head for hazard estimation.}

\subsection{Hierarchical WSI Encoding}
\label{sec:wsi}

\rev{Given a WSI $\mathbf{I}$, the hierarchical WSI encoding module retains the spatial
arrangement of patches and learns morphological representations across multiple scales.
Following the CLAM preprocessing workflow~\citep{lu2021}, we perform tissue segmentation
and extract $P$ non-overlapping $256 \times 256$-pixel patches
$\{(\mathbf{I}_p,\mathbf{s}_p)\}_{p=1}^{P}$, where $\mathbf{I}_p$ denotes the $p$-th patch
and $\mathbf{s}_p \in \mathbb{R}^{2}$ denotes its spatial coordinate. Pretrained feature
encoders map patches to $d_c$-dimensional embeddings, forming $\mathbf{X}^{\mathrm{patch}}
\in \mathbb{R}^{P \times d_c}$. Before spatial reconstruction, a learned channel
projection maps each patch embedding to $d_s=64$ channels. The projected embeddings are
arranged according to their coordinates to form a feature grid
$\mathbf{F}\in\mathbb{R}^{d_s\times H\times W}$, where $H$ and $W$ denote the grid height
and width. A binary mask $\Omega \in \{0,1\}^{H \times W}$ identifies locations occupied
by tissue patches, while unoccupied locations are zero-filled.}

\rev{Histopathological morphology spans multiple spatial scales, from local structures to
broader tissue organization~\citep{schmitz2021multiscale,vanrijthoven2021hooknet}. To
capture this hierarchy, we apply parallel $1\times1$, $3\times3$, and $5\times5$
convolutional branches to the reconstructed patch grid. Given the feature grid
$\mathbf{F}$ and tissue mask $\Omega$, the branch operations~are:}

\begin{equation}
  \label{eq:multi_conv}
  \begin{array}{@{}l@{\;}c@{\;}l@{\quad}l@{}}
    \widetilde{\mathbf{F}}_{(1)}
    & = & \displaystyle \operatorname{Conv}_{(1)}(\mathbf{F}),
    & \Omega_{(1)}=\Omega, \\ \noalign{\vskip 3pt}
    \left(\widetilde{\mathbf{F}}_{(l)},\Omega_{(l)}\right)
    & = & \displaystyle \operatorname{MPool}_{(l)}
          \left(\operatorname{Conv}_{(l)}(\mathbf{F}),\Omega\right),
    & l\in\{2,3\}, \\ \noalign{\vskip 3pt}
    \mathbf{X}^{\mathrm{wsi}}_{(l)}
    & = & \displaystyle \operatorname{Proj}_{\mathrm{out}}
          \left[\operatorname{Stack}
          \left(\widetilde{\mathbf{F}}_{(l)},\Omega_{(l)}\right)\right],
    & l\in\{1,2,3\}.
  \end{array}
\end{equation}

\rev{Here, $\operatorname{Conv}_{(l)}$ and $\operatorname{MPool}_{(l)}$ denote the $l$-th
convolutional branch and mask-aware average pooling, respectively, while $\Omega_{(l)}$ is
the corresponding valid-region mask. The operator $\operatorname{Stack}$ stacks valid
spatial feature vectors as matrix rows, and $\operatorname{Proj}_{\mathrm{out}}$ maps each
row to the shared dimension $d$. The resulting scale-specific WSI feature matrices
$\mathbf{X}^{\mathrm{wsi}}_{(l)}\in\mathbb{R}^{P_l\times d}$ are retained for
pathology--genomics interaction, where $P_l$ denotes the number of valid feature vectors
at scale $l$.}

\subsection{Scale-Wise Pathology--Genomics Fusion}

\rev{Given the WSI feature matrices at three scales,
$\{\mathbf{X}^{\mathrm{wsi}}_{(l)}\}_{l=1}^{3}$, HGASurv first constructs genomic
signature embeddings for scale-wise co-attention. Following the gene-signature
organization used in MCAT~\citep{chen2021}, genomic variables are grouped into six
gene-family signatures from the Molecular Signatures Database
(MSigDB)~\citep{subramanian2005}: Tumor Suppressor Genes; Cytokines and Growth Factors;
Oncogenes; Protein Kinases; Cell Differentiation Markers; and Transcription Factors. Each
signature-specific genomic profile is independently encoded by a self-normalizing network
(SNN)~\citep{klambauer2017} to produce a $d$-dimensional embedding. The six embeddings are
stacked into the genomic signature embedding matrix
$\hat{\mathbf{X}}^{\mathrm{gene}}\in\mathbb{R}^{6\times d}$, where $d$ is the shared
embedding dimension of the WSI and genomic representations.}

\rev{At each scale $l\in\{1,2,3\}$, the genomic signature embedding matrix provides the
queries, while the corresponding WSI feature matrix provides the keys and values:}

\begin{equation}
  \label{eq:cross_modal_attention}
  \begin{array}{@{}l@{\;}c@{\;}l@{}}
    \mathbf{Q}_r
    & = & \displaystyle \hat{\mathbf{X}}^{\mathrm{gene}}\mathbf{W}_r^Q, \\ \noalign{\vskip 3pt}
    \mathbf{K}_{r,(l)}
    & = & \displaystyle \mathbf{X}^{\mathrm{wsi}}_{(l)}\mathbf{W}_r^K, \\ \noalign{\vskip 3pt}
    \mathbf{V}_{r,(l)}
    & = & \displaystyle \mathbf{X}^{\mathrm{wsi}}_{(l)}\mathbf{W}_r^V, \\ \noalign{\vskip 3pt}
    \operatorname{head}_{r,(l)}
    & = & \displaystyle \operatorname{softmax}
          \left(\frac{\mathbf{Q}_r\mathbf{K}_{r,(l)}^{\top}}
          {\sqrt{d_a}}\right)\mathbf{V}_{r,(l)}, \\ \noalign{\vskip 3pt}
    \hat{\mathbf{X}}^{\mathrm{wsi}}_{(l)}
    & = & \displaystyle \operatorname{Concat}
          \left(\operatorname{head}_{1,(l)},\ldots,
          \operatorname{head}_{H_c,(l)}\right)\mathbf{W}^{O}.
  \end{array}
\end{equation}

\rev{Here, $\mathbf{W}_r^Q,\mathbf{W}_r^K,\mathbf{W}_r^V\in\mathbb{R}^{d\times d_a}$ and
$\mathbf{W}^{O}\in\mathbb{R}^{H_c d_a\times d}$ are learnable parameters, $H_c$ is the
number of co-attention heads, and $d_a=d/H_c$ is the dimension of each head. The softmax
is applied over the $P_l$ WSI feature vectors. The resulting matrix
$\hat{\mathbf{X}}^{\mathrm{wsi}}_{(l)}\in\mathbb{R}^{6\times d}$ contains one
signature-conditioned morphological representation for each genomic signature at scale
$l$.}

\rev{A single Transformer encoder~\citep{vaswani2017}, with parameters shared across the
three scales, is applied separately to each $\hat{\mathbf{X}}^{\mathrm{wsi}}_{(l)}$ to
model dependencies among its six signature-conditioned representations. Attention-based
pooling~\citep{ilse2018} then produces a scale-specific pathology representation
$\mathbf{Z}^{\mathrm{wsi}}_{(l)}\in\mathbb{R}^{d}$. The three scale-specific
representations are concatenated and fused by an MLP to obtain the final WSI
representation $\mathbf{Z}^{\mathrm{wsi}}\in\mathbb{R}^{d}$. In parallel, the genomic
signature embedding matrix is processed by a separate Transformer encoder and
attention-based pooling to produce $\mathbf{Z}^{\mathrm{gene}}\in\mathbb{R}^{d}$. Finally,
the WSI and genomic representations are concatenated to form the fused image--genomic
embedding
$\mathbf{Z}=[\mathbf{Z}^{\mathrm{wsi}};\mathbf{Z}^{\mathrm{gene}}]\in\mathbb{R}^{2d}$.}

\subsection{Clinical-Guided Graph Propagation}
\label{sec:clinical}

\rev{The preceding modules produce a fused image--genomic embedding
$\mathbf{Z}_i\in\mathbb{R}^{2d}$ for each patient $i$. Clinical variables define the
patient graph, while $\mathbf{Z}_i$ serves as the node feature for cross-patient
propagation.}

\noindent\textbf{Graph construction:} 
\rev{Because the clinical variables include both
continuous and categorical fields, we use Gower distance~\citep{gower1971} to measure
patient similarity. Let $d_{ij}\in[0,1]$ denote the distance between patients $i$ and $j$.
We define clinical similarity as $s_{ij}=1-d_{ij}$, where larger
values indicate greater similarity. We construct an undirected patient graph by
sparsifying the pairwise clinical-similarity matrix, with retained edges connecting
clinically similar patients.}

\noindent\textbf{Graph attention propagation:} 
\rev{Each fused embedding is projected to a
graph hidden dimension $d'$ and processed by $L_g$ residual graph attention layers with
$H_g$ heads~\citep{velickovic2018}. Let $\mathbf{u}_i^{(q)}\in\mathbb{R}^{d'}$ denote the
representation of patient $i$ after layer $q$, and let $\mathbf{a}_i^{(q)}$ denote the
neighborhood information aggregated at that layer. The graph propagation is formulated as}

\begin{equation}
  \label{eq:graph_attention}
  \begin{array}{@{}l@{\;}c@{\;}l@{}}
    \mathbf{u}_i^{(0)}
    & = & \displaystyle \mathbf{W}_0\mathbf{Z}_i+\mathbf{b}_0, \\ \noalign{\vskip 3pt}
    \mathbf{a}_i^{(q)}
    & = & \displaystyle \operatorname{AGG}_{k=1}^{H_g}
          \left[\sum_{j\in\mathcal{N}(i)}
          \alpha_{ij}^{(q,k)}\mathbf{W}^{(q,k)}
          \mathbf{u}_j^{(q-1)}\right], \\ \noalign{\vskip 3pt}
    \mathbf{u}_i^{(q)}
    & = & \displaystyle \operatorname{ELU}
          \left(\operatorname{LN}
          \left[\mathbf{u}_i^{(q-1)}+\mathbf{a}_i^{(q)}\right]\right), \\ \noalign{\vskip 3pt}
    \mathbf{z}'_i
    & = & \displaystyle \mathbf{u}_i^{(L_g)}.
  \end{array}
\end{equation}

\rev{Here, $\mathbf{W}_0$ and $\mathbf{b}_0$ define the initial projection, while
$\mathbf{W}^{(q,k)}$ is the feature projection for head $k$ in layer $q$. The neighborhood
$\mathcal{N}(i)$ includes patient $i$, and $\alpha_{ij}^{(q,k)}$ is the normalized
attention weight assigned to neighbor $j$. In intermediate layers, $\operatorname{AGG}$
concatenates $H_g$ head outputs of dimension $d'/H_g$; in the final layer, it averages
$H_g$ head outputs of dimension $d'$. The operators $\operatorname{LN}$ and
$\operatorname{ELU}$ denote layer normalization and ELU activation, respectively. The
output $\mathbf{z}'_i\in\mathbb{R}^{d'}$ is the graph-refined patient representation.}

\noindent\textbf{Survival prediction:} 
\rev{Finally, the graph-refined representation
$\mathbf{z}'_i$ is passed to the discrete-time survival head to estimate the
interval-specific hazards $\hat{h}_{i,t}$.}

\subsection{Training Objective}

\rev{For the $T$ intervals $\{\mathcal{B}_t\}_{t=1}^T$, let $b_i\in\{1,\ldots,T\}$ denote
the observed time bin of patient $i$, such that $t_i\in\mathcal{B}_{b_i}$. HGASurv
predicts the hazard probability $\hat{h}_{i,t}$ for each interval and is trained using the
negative log-likelihood (NLL) loss~\citep{zadeh2020}:}

\begin{equation}
  \begin{aligned}
    \mathcal{L}_{\mathrm{NLL}}
    &= -\frac{1}{N}\sum_{i=1}^{N}\Bigl[
       \sum_{t=1}^{b_i-1}\log(1-\hat{h}_{i,t}) \\
    &\quad+\delta_i\log(\hat{h}_{i,b_i})
       +(1-\delta_i)\log(1-\hat{h}_{i,b_i})\Bigr].
  \end{aligned}
\end{equation}

\rev{Here, $\delta_i=1$ denotes an observed event and $\delta_i=0$ denotes
right-censoring. The first sum represents survival before the observed bin, while the
remaining terms specify an event within bin $b_i$ or survival through that bin for event
and censored cases, respectively.}

\rev{HGASurv is trained in two stages to preserve the image--genomic representations
during graph integration. In Stage~1, the hierarchical WSI encoding, genomic encoding, and
cross-modal fusion modules are jointly trained with an MLP survival head by minimizing
$\mathcal{L}_{\mathrm{NLL}}$ using Adam. In Stage~2, the resulting image--genomic
embeddings are frozen, and only the graph propagation network and its survival head are
trained using the same NLL objective.\looseness=-1}

\section{Experiments}
\begin{table}[!htbp]
  \centering
  \caption{\rev{C-index comparison of HGASurv with state-of-the-art deep learning methods
  on three TCGA cohorts. Values are the mean $\pm$ sample standard deviation across five
  folds. The best result is shown in bold and the second-best is underlined. P, G, and C
  denote pathology, genomics, and clinical variables,~respectively.}}
  \label{tab:main_results}
  \linespread{1}\footnotesize
  \setlength{\tabcolsep}{3pt}
  \begin{tabularx}{\textwidth}{@{}>{\raggedright\arraybackslash}Xccccc@{}}
    \toprule
    Method & Year & Modality & TCGA-GBMLGG & TCGA-BLCA & TCGA-UCEC \\
    \midrule
    SNN \citep{klambauer2017} & 2017 & Genomic & $0.805 \pm 0.013$ & $0.567 \pm 0.067$ &
    $0.627 \pm 0.054$\\
    SNNTrans~\citep{klambauer2017} & 2017 & Genomic & $0.804 \pm 0.039$ & $0.630 \pm 0.021$
    & $0.621 \pm 0.056$\\
    AttnMIL \citep{ilse2018} & 2018 & Pathology & $0.759 \pm 0.040$ & $0.523 \pm 0.043$ &
    $0.665 \pm 0.029$\\
    DeepAttnMISL \citep{yao2020} & 2020 & Pathology & $0.736 \pm 0.036$ & $0.573 \pm 0.034$
    & $0.694 \pm 0.075$\\
    CLAM\_SB \citep{lu2021} & 2021 & Pathology & $0.753 \pm 0.041$ & $0.543 \pm 0.040$ &
    $0.674 \pm 0.057$\\
    CLAM\_MB \citep{lu2021} & 2021 & Pathology & $0.752 \pm 0.043$ & $0.542 \pm 0.021$ &
    $0.655 \pm 0.045$\\
    TransMIL \citep{shao2021} & 2021 & Pathology & $0.749 \pm 0.046$ & $0.523 \pm 0.039$ &
    $0.661 \pm 0.067$\\
    DTFD-MIL \citep{zhang2022} & 2022 & Pathology & $0.751 \pm 0.050$ & $0.533 \pm 0.065$ &
    $0.683 \pm 0.073$\\
    \midrule
    MCAT \citep{chen2021} & 2021 & P+G & $0.809 \pm 0.019$ & $0.632 \pm 0.036$ & $0.591 \pm
    0.027$\\
    Pathomic Fusion \citep{chen2020} & 2022 & P+G & $0.792 \pm 0.034$ & $0.611 \pm 0.046$ &
    $0.581 \pm 0.024$\\
    PORPOISE \citep{chen2022} & 2022 & P+G & $0.810 \pm 0.024$ & $0.605 \pm 0.046$ & $0.622
    \pm 0.058$\\
    MOTCat \citep{xu2023} & 2023 & P+G & $0.820 \pm 0.027$ & \underline{$0.684 \pm 0.026$}
    & $0.676 \pm 0.039$\\
    CMTA \citep{zhou2023cmta} & 2023 & P+G & $0.837 \pm 0.022$ & $0.663 \pm 0.036$ & $0.665
    \pm 0.065$\\
    MGCT \citep{liu2023} & 2023 & P+G & $0.829 \pm 0.022$ & $0.638 \pm 0.041$ & $0.647 \pm
    0.037$\\
    PG-MLIF \citep{pan2024pgmlif} & 2024 & P+G & $0.799 \pm 0.046$ & $0.595 \pm 0.019$ &
    $0.530 \pm 0.080$\\
    DeReF \citep{wang2025} & 2026 & P+G & \underline{$0.842 \pm 0.015$} & $0.666 \pm 0.026$
    & $0.666 \pm 0.042$\\
    SurMoE \citep{zhang2025surmoe} & 2026 & P+G & $0.832 \pm 0.013$ & $0.647 \pm 0.015$ &
    $0.669 \pm 0.042$\\
    \midrule
    \citet{sucre2025} & 2025 & P+G+C & $0.827 \pm 0.026$ & $0.590 \pm 0.045$ &
    $\mathbf{0.756 \pm 0.035}$\\
    SurvPGC \citep{hou2026survpgc} & 2026 & P+G+C & $0.823 \pm 0.019$ & $0.574 \pm 0.025$ &
    $0.701 \pm 0.070$\\
    \midrule
    \textbf{\rev{HGASurv (Ours)}} & 2026 & \textbf{P+G+C} & \rev{$\mathbf{0.854 \pm
    0.023}$} & \rev{$\mathbf{0.692 \pm 0.032}$} & \rev{\underline{$0.721 \pm 0.098$}} \\
    \bottomrule
  \end{tabularx}
\end{table}

\subsection{Datasets and Evaluation Protocol}

\rev{We evaluate HGASurv on three cancer cohorts from The Cancer Genome Atlas
(TCGA)~\citep{weinstein2013}: the combined glioblastoma and lower-grade glioma cohort
(GBMLGG), bladder urothelial carcinoma (BLCA), and uterine corpus endometrial carcinoma
(UCEC). The underlying whole-slide images, molecular profiles, and clinical data are
available through the Genomic Data Commons Data Portal
(\url{https://portal.gdc.cancer.gov/}) and cBioPortal
(\url{https://www.cbioportal.org/}), subject to the access conditions of those resources.
After matching patients with a diagnostic WSI, genomic profile, structured
clinical variables, and survival outcome, the resulting cohorts contain 551, 373, and 480
patients, respectively.}

\rev{All experiments use event-status-stratified, patient-level five-fold
cross-validation, with identical splits across methods within each cohort. We evaluate
survival prediction using the concordance index (C-index)~\citep{harrell1982}, with higher
values indicating better performance. Within each fold, each method is trained ten times
with independent random initializations. To align the scales of risk scores across these
runs before averaging, held-out scores from each run are rank-normalized using an
empirical cumulative distribution function fitted to that run's training-fold scores. The
ten normalized scores are then averaged to obtain one ensemble risk score for each
held-out patient, from which a fold-level C-index is computed.
Table~\ref{tab:main_results} reports the mean and sample standard deviation of the five
fold-level C-indices.}

\subsection{Comparison with State-of-the-Art Methods}

\rev{To benchmark HGASurv against state-of-the-art survival prediction methods, Table~\ref{tab:main_results} compares it with 19 baselines across the GBMLGG, BLCA, and UCEC cohorts. These methods use pathology images, genomic profiles, clinical variables, or their combinations.}
\rev{The genomic-only baselines comprise SNN~\citep{klambauer2017} and its self-attention-enhanced variant SNNTrans. The pathology-only group includes attention-based MIL methods (AttnMIL~\citep{ilse2018}, DeepAttnMISL~\citep{yao2020}, CLAM-SB, and CLAM-MB~\citep{lu2021}), Transformer-based aggregation (TransMIL~\citep{shao2021}), and feature distillation (DTFD-MIL~\citep{zhang2022}). }

\rev{The pathology--genomics baselines span three fusion strategies: feature-level integration with Pathomic Fusion~\citep{chen2020}, PORPOISE~\citep{chen2022}, and PG-MLIF~\citep{pan2024pgmlif}; cross-modal interaction and alignment with MCAT~\citep{chen2021}, MOTCat~\citep{xu2023}, CMTA~\citep{zhou2023cmta}, and MGCT~\citep{liu2023}; and adaptive expert fusion with DeReF~\citep{wang2025} and SurMoE~\citep{zhang2025surmoe}. The trimodal baselines include the prediction-level late-fusion approach of \citet{sucre2025} and the clinical-text-based cross-modal integration of SurvPGC~\citep{hou2026survpgc}.}

\rev{The best multimodal result exceeds the best unimodal result in each cohort. However, the trimodal baselines outperform the strongest pathology--genomics baseline on UCEC but not on GBMLGG or BLCA, suggesting that adding clinical information does not guarantee consistent improvements across integration strategies and cohorts.}

\begin{figure}[!htbp]
\centering
\includegraphics[width=\textwidth]{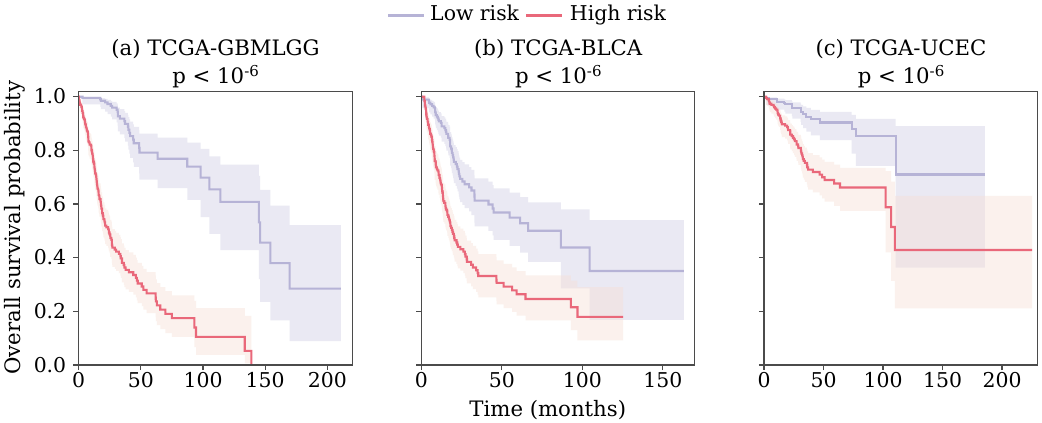}
\caption{\rev{Kaplan--Meier survival curves stratified by the final HGASurv out-of-fold risk predictions on (a) TCGA-GBMLGG, (b) TCGA-BLCA, and (c) TCGA-UCEC.}}
\label{fig:km}
\end{figure}

\rev{HGASurv achieves mean C-indices of 0.854, 0.692, and 0.721 on GBMLGG, BLCA, and UCEC, respectively, ranking first on GBMLGG and BLCA and second on UCEC, thereby demonstrating consistent performance across all three cohorts. HGASurv combines within-patient pathology--genomics interactions with clinically defined relationships between patients; the component and clinical-integration ablations in Tables~\ref{tab:ablation} and~\ref{tab:clinical_integration} further examine the contributions of these design elements. The out-of-fold risk predictions stratify patients into groups with significantly different survival outcomes across all three cohorts as Fig.~\ref{fig:km} shows.}

\subsection{Ablation Studies}
\subsubsection{Component Ablation}

\rev{To isolate the individual and joint contributions of hierarchical WSI encoding and clinical-guided graph propagation, Table~\ref{tab:ablation} compares the full HGASurv model with three variants that remove graph propagation, replace hierarchical WSI encoding with the single $1\times1$ pathology branch, or apply both changes.}

\begin{table}[!htbp]
  \centering
  \caption{\rev{Component ablation of HGASurv across three TCGA cohorts.}}
  \label{tab:ablation}
  \setlength{\tabcolsep}{4pt}
  \renewcommand{\arraystretch}{1.3}
  \small
  \newcommand{\ablationcfg}[1]{\parbox[t]{0.25\textwidth}{\raggedright\linespread{1}\selectfont #1}}
  \begin{tabular}{@{}lccccc@{}}
    \toprule
    Configuration & \shortstack{Hier.\\WSI} & \shortstack{Clinical\\Variables} & GBMLGG &
    BLCA & UCEC \\
    \midrule
    \ablationcfg{Full HGASurv} & \checkmark & \checkmark & $0.854\!\pm\!0.023$ &
    $0.692\!\pm\!0.032$ & $0.721\!\pm\!0.098$ \\\addlinespace[5pt]
    \ablationcfg{w/o Clinical Variables} & \checkmark & $\times$ & $0.803\!\pm\!0.045$ &
    $0.655\!\pm\!0.037$ & $0.616\!\pm\!0.076$ \\\addlinespace[5pt]
    \ablationcfg{w/o Hierarchical WSI} & $\times$ & \checkmark & $0.806\!\pm\!0.032$ &
    $0.566\!\pm\!0.027$ & $0.660\!\pm\!0.072$ \\\addlinespace[5pt]
    \ablationcfg{w/o both modules} & $\times$ & $\times$ & $0.787\!\pm\!0.034$ &
    $0.587\!\pm\!0.036$ & $0.597\!\pm\!0.107$ \\\addlinespace[5pt]
    \bottomrule
  \end{tabular}
\end{table}

\rev{Removing either hierarchical WSI encoding or clinical-guided graph propagation reduces the mean C-index across all three cohorts (Table~\ref{tab:ablation}). The ablated variants fall below the strongest pathology–genomics baseline in each cohort (Table~\ref{tab:main_results}), whereas full HGASurv outperforms these baselines. Removing both components further degrades performance on GBMLGG and UCEC. These findings indicate that the two components provide complementary benefits and are key contributors to HGASurv’s performance advantage over pathology–genomics baselines.}

\subsubsection{Clinical Integration Strategies}

\rev{To assess whether clinical variables improve survival prediction and how they should be incorporated, Table~\ref{tab:clinical_integration} compares three configurations while keeping the image--genomic encoder, patient splits, and training and evaluation protocols fixed. First, the \emph{No clinical variables} configuration predicts survival directly from the image--genomic embedding without clinical inputs or graph propagation. Second, \emph{Clinical feature concatenation} combines encoded clinical variables with the image--genomic embedding before the MLP survival head. Third, the proposed \emph{Clinical-guided graph propagation} configuration uses clinical variables to construct a patient graph and propagates image--genomic embeddings between clinically similar patients before survival prediction.}

\begin{table}[!htbp]
\centering
\caption{\rev{C-index comparison of clinical integration configurations. Values are the mean $\pm$ sample standard deviation across five folds. The best result in each cohort is shown in bold.}}
\label{tab:clinical_integration}
\setlength{\tabcolsep}{6pt}
\renewcommand{\arraystretch}{1.3}
\small
\newcommand{\clinicalcfg}[1]{\parbox[t]{0.30\textwidth}{\raggedright\linespread{1}\selectfont #1}}
\begin{tabular}{@{}lccc@{}}
\toprule
Clinical integration & GBMLGG & BLCA & UCEC \\\addlinespace[5pt]
\midrule
\clinicalcfg{No clinical variables} & $0.803 \pm 0.045$ & $0.655 \pm 0.037$ & $0.616 \pm 0.076$ \\\addlinespace[5pt]
\clinicalcfg{Clinical feature concatenation} & $0.824 \pm 0.046$ & $0.664 \pm 0.026$ & $0.666 \pm 0.102$ \\\addlinespace[5pt]
\clinicalcfg{\textbf{Clinical-guided graph propagation~(Ours)}} & $\mathbf{0.854 \pm 0.023}$ & $\mathbf{0.692 \pm 0.032}$ & $\mathbf{0.721 \pm 0.098}$ \\\addlinespace[5pt]
\bottomrule
\end{tabular}
\end{table}

\rev{Both clinical feature concatenation and clinical-guided graph propagation improve the mean C-index over the no-clinical configuration in all three cohorts, with graph propagation performing best. These results suggest that clinical variables are more effective for guiding cross-patient propagation than for direct feature concatenation in these cohorts.}

\begin{figure}[!htbp]
  \centering
  \includegraphics[width=\textwidth]{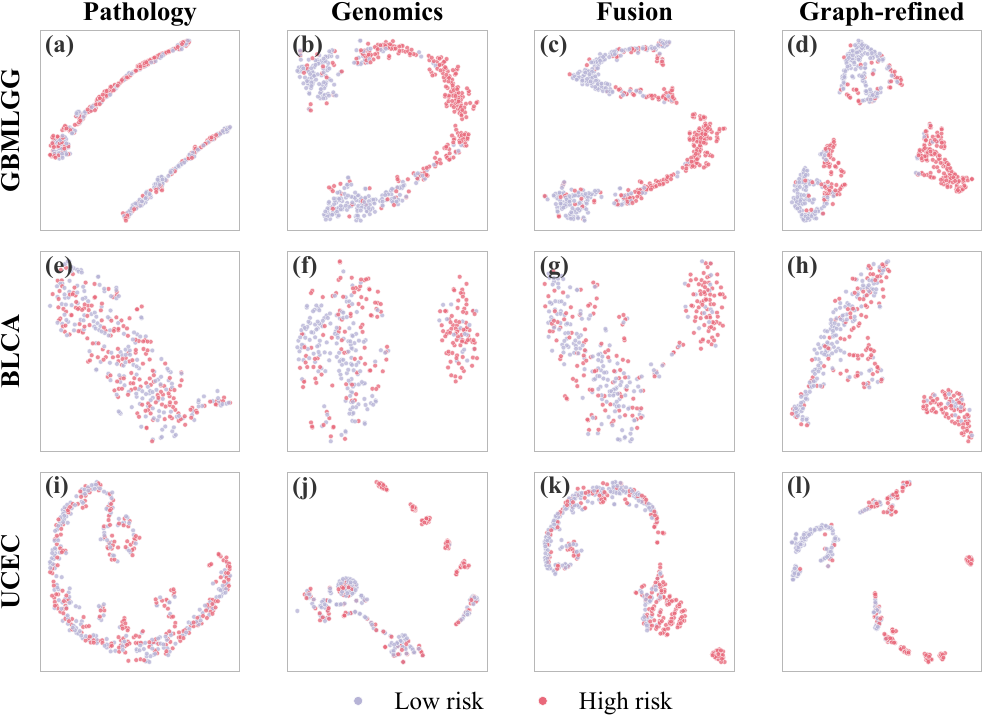}
  \caption{\rev{t-distributed stochastic neighbor embedding (t-SNE) of pathology, genomic, fused image--genomic, and clinical-graph-refined representations in TCGA-GBMLGG, TCGA-BLCA, and TCGA-UCEC (top to bottom). Colors denote predicted high- and low-risk groups.}}
  \label{fig:tsne}
\end{figure}

\subsubsection{Graph Topology Controls}

\rev{To assess the contribution of clinically informed patient connectivity, Table~\ref{tab:topology} compares three graph topologies. The \emph{Self-loop graph} retains the GAT and survival head with only self-connections to evaluate graph processing without cross-patient information exchange. The \emph{Random graph} uses Chung--Lu sampling~\citep{chung2002} based on the clinical graph's node degrees to test whether arbitrary patient connections yield comparable benefits. The \emph{Clinical graph} connects patients according to clinical similarity to assess the benefit of clinically guided information exchange.}

\begin{table}[!htbp]
\centering
\caption{\rev{C-index comparison of patient-graph topology controls. Values are the mean $\pm$ sample standard deviation across five folds. The best result in each cohort is shown in bold.}}
\label{tab:topology}
\setlength{\tabcolsep}{6pt}
\renewcommand{\arraystretch}{1.3}
\small
\newcommand{\topologycfg}[1]{\parbox[t]{0.30\textwidth}{\raggedright\linespread{1}\selectfont #1}}
\begin{tabular}{@{}lccc@{}}
\toprule
Graph topology & GBMLGG & BLCA & UCEC \\\addlinespace[5pt]
\midrule
\topologycfg{Self-loop graph} & $0.822 \pm 0.034$ & $0.681 \pm 0.029$ & $0.681 \pm 0.071$ \\\addlinespace[5pt]
\topologycfg{Random graph} & $0.820 \pm 0.042$ & $0.681 \pm 0.032$ & $0.666 \pm 0.067$ \\\addlinespace[5pt]
\topologycfg{Clinical graph~(Ours)} & $\mathbf{0.854 \pm 0.023}$ & $\mathbf{0.692 \pm 0.032}$ & $\mathbf{0.721 \pm 0.098}$ \\\addlinespace[5pt]
\bottomrule
\end{tabular}
\end{table}

\rev{The random graph does not improve upon the self-loop control in any cohort, indicating that arbitrary cross-patient connectivity is insufficient. In contrast, the clinical graph achieves the highest mean C-index in all three cohorts. These results highlight the importance of clinically meaningful topology for cross-patient modeling.}

\subsection{\rev{Visualization and Analysis}}

\subsubsection{Survival Risk Stratification}

\rev{To visualize HGASurv's risk stratification, Fig.~\ref{fig:km} shows Kaplan--Meier curves stratified by each cohort's median out-of-fold predicted risk. The high-risk group shows lower survival across all three cohorts (log-rank $p<10^{-6}$).}

\subsubsection{Patient Representation Visualization}

\rev{To visualize how fusion and clinical-guided graph propagation organize patient representations, Fig.~\ref{fig:tsne} shows t-SNE projections of pathology, genomic, fused image--genomic, and graph-refined representations. Each point represents a patient, with the same final predicted high-/low-risk coloring across each cohort's four panels. Risk groups overlap before fusion; after fusion and graph propagation, regions dominated by different groups are visible.}

\section{Conclusion}
\rev{We presented HGASurv, a multimodal survival prediction framework that represents and
integrates pathology, genomics, and clinical variables according to their information
granularity. Within pathology, patch features retain their spatial arrangement and are
encoded at patch, local-neighborhood, and regional scales; across modalities,
low-dimensional clinical variables shape predictions through the connectivity of a
patient-similarity graph rather than competing with high-dimensional image–genomic
embeddings as concatenated features. On three TCGA cohorts, HGASurv achieved the strongest
overall C-index performance among the evaluated methods, and component ablations supported
the complementary contributions of hierarchical WSI encoding and graph-based clinical
integration. Moreover, graph-based clinical integration outperformed direct concatenation
of the same clinical variables in all three cohorts.}

\rev{Although these results suggest a benefit of tailoring representation and integration
to modality-specific information granularity, further investigation is needed to establish
how far this benefit extends beyond the present evaluation and how it arises in individual
predictions. Future work will evaluate HGASurv in cohorts independent of TCGA and assess
its sensitivity to feature-extraction and graph-construction choices.  Patient-level
analyses will investigate how scale-specific pathology representations and information
exchanged through clinical neighborhoods shape individual predictions.}

\FloatBarrier
\section*{Funding}
This work was supported in part by the National Natural Science Foundation of China
[grant numbers 62372326, 62172300]; the Shanghai Municipal Education Commission
Artificial Intelligence-Driven Research Paradigm Reform and Discipline Leapfrogging
Empowerment Program [grant number 2024RGYB001]; and the China Postdoctoral Science
Foundation [grant number 2025M771493].

\end{document}